\documentclass[letterpaper, 10pt, journal, twoside]{IEEEtran}

\IEEEoverridecommandlockouts  

\usepackage[pdftex]{graphicx}
\usepackage{epsfig} 
\usepackage{times} 
\usepackage{amsmath} 
\usepackage{amssymb}  
\usepackage{url}
\usepackage{hyperref}

\usepackage{comment}
\usepackage{mathtools}
\usepackage[font=footnotesize]{caption}
\usepackage[subrefformat=parens,font=footnotesize]{subcaption}
\usepackage[utf8]{inputenc}

\usepackage[ruled,linesnumbered,noend]{algorithm2e}
\SetAlCapFnt{\footnotesize}
\usepackage{footmisc}

\makeatletter
\newcommand{\figcaption}[1]{\def\@captype{figure}\caption{#1}}
\newcommand{\tblcaption}[1]{\def\@captype{table}\caption{#1}}
\makeatother

\newcommand{\argmin}{\mathop{\rm arg~min}\limits}
\newcommand{\equationname}{Eq.~}
\newcommand{\tablenameref}{TABLE~}

\newcommand{\eqrefs}[2]{Eqs. \ref{#1}--\ref{#2}}

\usepackage{booktabs}
\usepackage{longtable}
\usepackage{tabularx}
\usepackage{makecell}
\usepackage{color}

\title{\LARGE \bf Cutting Sequence Diffuser: Sim-to-Real Transferable Planning \\ for Object Shaping by Grinding}

\author{Takumi Hachimine$^{1}$, Jun Morimoto$^{2,3}$, and Takamitsu Matsubara$^{1}$
\thanks{This work was supported by JST-Mirai Program Grant Number~JPMJMI21B1, Japan.}
\thanks{$^{1}$T. Hachimine~and~T. Matsubara are with
the Graduate School of Information Science, Nara Institute of Science and
Technology (NAIST), Nara, Japan.}%
\thanks{$^{2}$J. Morimoto is with the Department of Systems Science, Graduate
School of Informatics, Kyoto University, Kyoto, Japan.}%
\thanks{$^{3}$J. Morimoto is also with the Brain Information Communication Research Laboratory Group (BICR), Advanced Telecommunications Research
Institute International (ATR), Kyoto, Japan.}%
}

\begin{document}
\maketitle
\thispagestyle{empty}
\pagestyle{empty}

\begin{abstract}
Automating object shaping by grinding with a robot is a crucial industrial process that involves removing material with a rotating grinding belt. This process generates removal resistance depending on such process conditions as material type, removal volume, and robot grinding posture, all of which complicate the analytical modeling of shape transitions. Additionally, a data-driven approach based on real-world data is challenging due to high data collection costs and the irreversible nature of the process. This paper proposes a Cutting Sequence Diffuser (CSD) for object shaping by grinding. The CSD, which only requires simple simulation data for model learning, offers an efficient way to plan long-horizon action sequences transferable to the real world. Our method designs a smooth action space with constrained small removal volumes to suppress the complexity of the shape transitions caused by removal resistance, thus reducing the reality gap in simulations. Moreover, by using a diffusion model to generate long-horizon action sequences, our approach reduces the planning time and allows for grinding the target shape while adhering to the constraints of a small removal volume per step. Through evaluations in both simulation and real robot experiments, we confirmed that our CSD was effective for grinding to different materials and various target shapes in a short time.
\end{abstract}


\section{INTRODUCTION}
\label{section:introduction}
Object-shape manipulation is an important skill in industry and our daily lives \cite{arriola2020modeling}.
A shaping method that gradually removes unnecessary material from a base stock is called a removal process.
As a common industrial removal process, this paper focuses on the grinding process in which a rotating grinding belt removes material and explores the automation of this process with a robot.

The grinding process generates removal resistance depending on the process conditions (e.g., material type, removal volume, robot grinding posture, etc.). 
Such removal resistance drags the robot's end-effector and leads to shaping errors, complicating the analytical modeling of the objects' shape transitions.
Therefore, modeling shape transitions with a data-driven approach using real-world data may be reasonable.
In particular, object-shaping by Model-Based Reinforcement Learning (MBRL) has shown its effectiveness by iteratively optimizing both shape transition models by learning and action plans using Model Predictive Control (MPC) \cite{10214100}.
However, due to the irreversible nature of the removal process, the high cost of data collection remains a severe issue, emphasizing the importance of addressing this problem from a practical standpoint.

\begin{figure}[t]
    \centering
    \includegraphics[clip,width=1.0\linewidth]{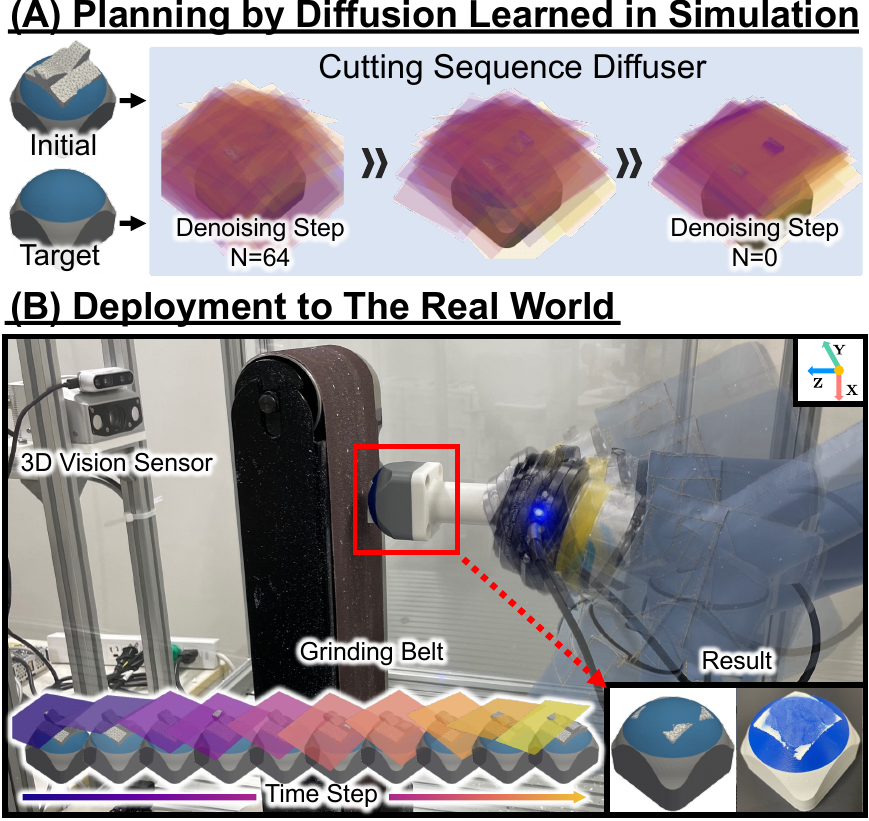}
    \caption{
    Trajectory generation for object shaping by grinding. The proposed method (cutting sequence diffuser) enables sim-to-real transferable trajectory generation using a diffusion model by constraining the robot's action with a small removal volume, which reduces removal resistance.}
    \label{fig:eye_cath}
    \vspace{-4truemm}
\end{figure}

This study utilizes simulations to explore an action-planning method that requires no data collection in the real world.
Ignoring the removal resistance, the shape transition of objects due to grinding can be viewed as a geometric splitting of the shape with a cutting surface (belt) and simulated by algebraic operations. 
However, removal resistance creates a reality gap between the real world and simulations. 

The grinding resistance theory \cite{TANG20092847}, which states that removal resistance is proportional to the removal volume of one step, suggests that reducing the removal volume in the grinding action decreases resistance and thus the reality gap. 
This approach should enable the direct transfer of simulated action sequences to the real world. However, such an approach lengthens task steps and increases computational costs, necessitating a long-horizon action sequence planner. 
Recent generative models, such as diffusion models \cite{sohl2015deep,ho2020denoising, NEURIPS2021_49ad23d1}, have shown the ability to learn long-horizon trajectory distributions \cite{carvalho2023motion,janner2022diffuser}. This suggests that sampling from these distributions could serve as an effective long-horizon action sequence planner.

With the above in mind, we propose a Cutting Sequence Diffuser (CSD) for object shaping by grinding (\figurename\ref{fig:eye_cath}).
Our method constrains the robot's action with a small removal volume, which reduces removal resistance to simplify the shape transition as a geometric cutting model. This transition model is then applied to data collection for an action sequence planner by a diffusion model.
Diffusion models can represent flexible distributions because data distributions are learned by gradually removing noise from Gaussian noise \cite{ho2020denoising}.
Therefore, sampling from a distribution conditioned on the cost of the target behavior (trajectory) enables long-horizon action sequences to be planned quickly, which has been difficult to accomplish with conventional optimization-based action planning methods \cite{carvalho2023motion,janner2022diffuser}. 

We evaluate the proposed method in both simulation and real robot experiments. These results indicate that our method enables a sim-to-real transfer due to action constraints that reduce the removal resistance. Moreover, an action plan by a diffusion model quickly provides flexible long-horizon action sequences for different materials and various target shapes.

The following are the main contributions of this paper:
\begin{itemize}
    \item We propose a sim-to-real transferable shape transition model that constrains a robot's action with a small removal volume to reduce removal resistance.
    \item We propose a novel cutting sequence planning framework, CSD, which utilizes a diffusion model for object shaping by grinding.
    \item We validate that CSD enables direct transfer of the planned action sequence to the real world through simulation and real robot experiments.
\end{itemize}

\section{RELATED WORK}
\subsection{Learning to Plan for Object-shape Manipulation}
Many studies have been conducted on object-shaping tasks with robots.
The reinforcement learning approach is practical because there is no risk of injury or physical burden on the human demonstrators compared to imitation learning which uses demonstration data.
The deformation process is a typical method of shaping objects by bending or pressing as if forming dough.
Since this process is reversible and can revert to its original shape, data collection costs are low.
Some studies have been conducted using collected data from the real world.
Matl \textit{et al}. automated a dough-shaping task using MBRL \cite{matl2021deformable}.
In addition, since the deformation process does not generate removal resistance, learned policies can be easily transferred on a simulator to the real world \cite{liang2018gpu,huang2021plasticinelab}.

In contrast, the removal process is irreversible and cannot revert to its original shape, which causes high data collection costs.
For the grinding process, CSA-MBRL \cite{10214100} has been proposed to plan actions by learning a shape transition model that considers removal resistance using real data.
The finite element or particle methods are used as simulators for removal processes but still suffer from a reality gap due to removal resistance.
Another method has been proposed to identify the dynamics parameters of a simulator using real-world data \cite{beltran2024sliceit}. 
Heiden \textit{et al}. proposed a cutting simulator to replicate the force applied to a knife when cutting ingredients using real-world data \cite{heiden2021disect}.
However, such methods only identify the dynamics parameters of the simulator and do not plan robot actions for object shaping.

Thus, the automation of removal processes has been limited to approaches that either learn from real data or identify simulator parameters.
In contrast, the novelty of our proposed method is its ability to transfer the planned actions in a simulator to the real world.
In other words, our method does not require real data, unlike CSA-MBRL.

\subsection{Robot Control with Diffusion Models}
Diffusion models \cite{sohl2015deep,ho2020denoising,NEURIPS2021_49ad23d1} are a type of generative model known for their high expressiveness and flexibility. 
They generate data through a denoising process and can conduct conditional sampling via guide (cost) functions.
This capability allows for various applications, such as text-conditioned image generation \cite{NEURIPS2021_49ad23d1} and video editing \cite{pmlr-v235-cohen24a}.

As examples of applying diffusion models for robots, Mishra \textit{et al}. proposed a task and motion planning method that combines individual task skills \cite{mishra2023generative}.
In the context of action (motion) planning, Janner \textit{et al}. proposed a diffuser \cite{janner2022diffuser}. 
They show that long-horizon trajectories conditioned on such guide functions as rewards can be planned using a diffusion model. 
In addition, Chi \textit{et al}. proposed a visuomotor policy that plans robot actions from image inputs using a diffusion model \cite{chi2023diffusionpolicy, chi2024universal}. 
Learned policy has been validated across 11 tasks that require multi-modal action.
Diffusion models have also been applied to contact-rich manipulation tasks such as wiping \cite{okada2024contact}.
In another study, Urain \textit{et al}. proposed a SE(3) diffusion model for 6-DoF grasping that optimizes a robot's motion and grasping posture \cite{urain2023se}.

Our method follows the features of flexible planning for long-horizon action sequences in a short time compared to conventional optimization-based action planning methods. 
Consequently, we propose a novel object-shaping method that is the first to apply a diffusion model as an action planner for the grinding process.
Furthermore, through long-horizon action sequence planning using a diffusion model conditioned by guide (cost) functions, our method enables shaping different materials and various target shapes.

\section{Preliminary} 
In preparation for the proposed method, \S \ref{subsection:shape_transition_modeling_and_planning} describes shape transition modeling using a cutting surface, which can be constructed by constraining a robot's action to a small removal volume. Additionally, we formulate the action planning problem.
\S \ref{subsection:trajectory_generation_with_diffusion_model}, introduces a trajectory generation (action planning) method using a diffusion model \cite{carvalho2023motion,janner2022diffuser}.

\subsection{Object Shaping Planning for Grinding}\label{subsection:shape_transition_modeling_and_planning}
\subsubsection{Shape Transition Modeling by Cutting Surface}\label{subsubsection:shape_transition_modeling}
The grinding process is performed by local surface contact between the tool surface (belt) and the object shape.
Ignoring removal resistance, shape transition can be viewed as a geometric splitting by the cutting surface (\figurename\ref{fig:cutting_surface_move}, top).
Here, the robot's action is equivalent to specifying the cutting surface.
This step allows us to represent the shape transition model as a Geometric-Cutting Model (GCM), where cutting surface (robot action) $\mathbf{a}_{t}\in \mathcal{A}$ geometrically splits current shape $\mathbf{s}_{t}\in \mathcal{S}$ into next time-step shape $\mathbf{s}_{t+1}$ and removal shape $\mathbf{w}_{t+1}$.
The shape transition by the cutting surface is defined as the following functions \cite{10214100}:
\begin{align}
    {\mathbf s}_{t+1}^{}= {\Psi_s}\left(\mathbf{s}_t,~{\mathbf{a}}_t\right), 
    {\mathbf w}_{t+1}^{}= {\Psi_w}\left(\mathbf{s}_t,~{\mathbf{a}}_t\right).
    \label{eq:cutting_function}
\end{align}
These functions split the shape using geometric collision detection, simplifying their implementation.
\figurename\ref{fig:cutting_surface_move} (bottom) shows shape deformation by cutting surface $\mathbf{a}_{t}$.

In the actual grinding process, removal resistance is generated depending on the process conditions.
The removal resistance generated by grinding is called grinding resistance \(F_t\), which is basically proportional to removal volume \(V_t\) and inversely proportional to belt rotation speed \(S_g\), \(F_t = \eta{{V}_{t}}/{S_g}\) \cite{TANG20092847}.
Here \(\eta\) is a coefficient, derived from the material type and the belt's grinding performance.
If the belt rotation speed is constant, the grinding resistance is proportional to the removal volume.
Thus, a cutting action with a large removal volume increases the grinding resistance and the actual shape transition contradicts the GCM.

\subsubsection{Cutting Sequence Optimization Problem}
Given a shape transition model, we denote trajectory $\boldsymbol{\tau}\coloneqq[\mathbf{s}_{t},\mathbf{a}_{t},\ldots,\mathbf{s}_{t+H-1},\mathbf{a}_{t+H-1}]$ and cost function $c(\boldsymbol{\tau})$, 
optimal robot action sequences $\mathbf{a}_{t:t+H-1}^{*}$ can be formulated as follows:
\begin{equation}
 \left.
 \begin{gathered}
\mathbf{a}_{t:t+H-1}^{*} = 
\argmin_{{{\mathbf a}_{t:t+H-1}  }}c(\boldsymbol{\tau}),\\
  \text{subject to}~\mathbf{s}_{t+1}={\Psi_s}\left(\mathbf{s}_{t},\mathbf{a}_t\right).
 \end{gathered}
 \right\}
\label{eq:mpc_objective_function}
\end{equation}
Here \(H\) is the planning horizon and \(t\) is the task index.
Deriving action sequences that satisfy \equationname\ref{eq:mpc_objective_function} using conventional optimization-based action planning methods (e.g., random shooting method \cite{nagabandi2018neural}), significantly increases the computational cost depending on the planning horizon.
Hence, long-horizon action planning is time-consuming \cite{carvalho2023motion,janner2022diffuser}.

\subsection{Trajectory Generation with Diffusion Models}\label{subsection:trajectory_generation_with_diffusion_model}
\subsubsection{Denoising Diffusion Probabilistic Models}
Our framework generates trajectory $\boldsymbol{\tau}$ using diffusion models.
They consist of a forward diffusion process that transforms the trajectory from data distribution into Gaussian noise and a denoising (inverse) process that transforms the Gaussian noise back to data distribution.
Given that original data $\boldsymbol{\tau}^{0}\equiv\boldsymbol{\tau}$ and $\boldsymbol{\tau}^{i}$ denote the data after adding $i$ times noise, the forward diffusion process is defined:
\(q(\boldsymbol{\tau}^{1:N}|\boldsymbol{\tau}^{0})\coloneqq {\prod_{i=1}^{N}} q(\boldsymbol{\tau}^{i}|\boldsymbol{\tau}^{i-1}), q(\boldsymbol{\tau}^{i}|\boldsymbol{\tau}^{i-1})\coloneqq\mathcal{N}(\boldsymbol{\tau}^{i}; \sqrt{{1-\beta_{i}}}\boldsymbol{\tau}^{i-1},\beta_{i}\mathbf{I}),\)
where $\beta_{i}$ is a noise scheduler that controls the noise scale.

The denoising process is parametrized as a Gaussian form, and the mean and covariance are defined by a model with parameter $\theta$ which takes noise data $\boldsymbol{\tau}^{i}$ and diffusion steps $i$ as input:
\begin{align}
p_{\theta}(\boldsymbol{\tau}^{0:N})&\coloneqq p(\boldsymbol{\tau}^{N}) {\prod_{i=1}^{N}}p_{\theta}(\boldsymbol{\tau}^{i-1}|\boldsymbol{\tau}^{i}),\\
p_{\theta}(\boldsymbol{\tau}^{i-1}|\boldsymbol{\tau}^i)&\coloneqq \mathcal{N}(\boldsymbol{\tau}^{i-1};\boldsymbol{\mu}_{\theta},(\boldsymbol{\tau}^{i},i),\boldsymbol{\Sigma}_{\theta}(\boldsymbol{\tau}^{i},i)) ,\label{eq:diffusion_reverse_process}\\
p_{\theta}(\boldsymbol{\tau}^{N})&=\mathcal{N}(\mathbf{0},\mathbf{I}).
\end{align}
Covariance $\boldsymbol{\Sigma}_{\theta}(\boldsymbol{\tau}^{i}, i)$ denotes $\boldsymbol{\Sigma}_{\theta}(\boldsymbol{\tau}^{i},i)=\sigma_{i}\mathbf{I}=\Tilde{\beta_{i}}\mathbf{I}$ and independent of parameter $\theta$. Where $\Tilde{\beta_{i}}=\beta_{i}(1-\bar{\alpha}_{i-1})/(1-\bar{\alpha}_{i})$, $\bar{\alpha_{i}}=\prod_{s=1}^{i}\alpha_{s}$, and \(\alpha_{t}=1-\beta_{t}\).
In the learning process, instead of directly optimizing $\boldsymbol{\mu}_{\theta}$, noise $\epsilon$ added to data $\boldsymbol{\tau}^{i}$, can be learned by simplified loss function 
$\mathcal{L}_{\theta}=\mathbb{E}_{i,\epsilon,\boldsymbol{\tau}^{0}}\left[\|\epsilon-\epsilon_{\theta}(\boldsymbol{\tau}^{i},i)	\|^2\right]$,
since $\boldsymbol{\mu}_{\theta}$ can be represented by $\epsilon_{\theta}$ as $\boldsymbol{\mu}_{\theta}(\boldsymbol{\tau}^{i},i)=\frac{1}{\sqrt{\alpha_{i}}}(\boldsymbol{\tau}^{i}-\frac{1-\alpha_{i}}{\sqrt{1-\alpha_{i}}}\epsilon_{\theta}(\boldsymbol{\tau}^{i},i))$ \cite{ho2020denoising}.

\subsubsection{Diffusion Models with Guided Sampling}
By introducing, a binary variable to indicate the trajectory as optimal by $\mathcal{O}=1$ and non-optimal by $\mathcal{O}=0$, and using the Control as Inference, trajectory optimization can be formulated as a stochastic inference problem that samples the optimal trajectory from distribution by
\(
\tilde{p}_{\theta}(\boldsymbol{\tau})=p(\boldsymbol{\tau}|\mathcal{O}=1)\propto p_{\theta}(\boldsymbol{\tau})p(\mathcal{O}=1|\boldsymbol{\tau}).
\)
Here $p_{\theta}(\boldsymbol{\tau}^{})=\int p(\boldsymbol{\tau}^{N}) \prod_{i=1}^{N}p_{\theta}(\boldsymbol{\tau}^{i-1}|\boldsymbol{\tau}^{i})d\boldsymbol{\tau}^{1:N}$ is a prior distribution of the trajectory, and $p(\mathcal{O}|\boldsymbol{\tau}^{})$ is the likelihood for the optimality of the trajectory. 
As a common assumption, since \(p\) belongs to an exponential family, we can arbitrarily write \(p(\mathcal{O}=1|\boldsymbol{\tau}^{})\propto \exp({-c_{}(\boldsymbol{\tau}))}\).

The posterior distribution of the denoising process can be expressed in Gaussian form for each denoising process $p(\boldsymbol{\tau}^{}|\mathcal{O}=1)=p_{\theta}(\boldsymbol{\tau}^{0:N})p(\mathcal{O}=1|\boldsymbol{\tau})$. Therefore, an approximate formulation is obtained as follows \cite{sohl2015deep, NEURIPS2021_49ad23d1}:
\begin{equation}
    {p}_{\theta}(\boldsymbol{\tau}^{i-1}|\boldsymbol{\tau}^i,\mathcal{O}={1})\approx \mathcal{N}(\boldsymbol{\tau}^{i-1};\boldsymbol{\mu}+{\boldsymbol{\Sigma}}g,\boldsymbol{\Sigma}),
\end{equation} 
where $\boldsymbol{\mu}$ and $\boldsymbol{\Sigma}$ are given by {\equationname\ref{eq:diffusion_reverse_process}:
$\boldsymbol{\mu}= \boldsymbol{\mu}_{\theta}(\boldsymbol{\tau}^{i},.i),\boldsymbol{\Sigma}=\boldsymbol{\Sigma}_{\theta}(\boldsymbol {\tau}^{i}, i)$. 
$g$ is derived from the gradient of the cost function:
\begin{align}
g&=\nabla_{{\boldsymbol{\tau}}^{i-1}}\log p\left(\mathcal{O}=1|\boldsymbol{\tau}^{i-1}\right)|_{\boldsymbol{\tau}^{i-1}=\boldsymbol{\mu}^{}}, \nonumber\\
&=-{\nabla_{\boldsymbol{\tau}^{i-1}}}c_{}\left(\boldsymbol{\tau}^{i-1}\right)|_{\boldsymbol{\tau}^{i-1}={\boldsymbol{\mu}^{}}_{}}.
\end{align}
Thus, by shifting mean $\boldsymbol{\mu}$ in the denoising process based on the gradient of a cost function designed for the task (e.g., state constraints, trajectory smoothness), we can generate trajectories guided by the cost function.

In addition, diffusion models are trained for the accuracy of the generated trajectories rather than single-step errors. 
Hence, they can quickly plan long-horizon action sequences without suffering from the significant increase in computational cost that is common in conventional optimization-based action planning methods \cite{carvalho2023motion,janner2022diffuser}.

\section{Proposed Method}
This section explains our planning method for cutting sequences using a diffusion model.
\S \ref{subsection:CSD_framework} shows the framework of our proposed method, CSD.
The following sections describe the components of the framework: \S \ref{subsection:action_constrain} describes the action constraints used for data collection, \S \ref{subsection:shape_encoding} describes a method for reducing high-dimensional shape information to lower dimensions, and \S\ref{subsection:diffusion_cost_design} describes the design and usage of the cost functions for guides during trajectory generation.
\begin{figure}[t!]
    \centering
    \includegraphics[clip,width=0.95\linewidth]{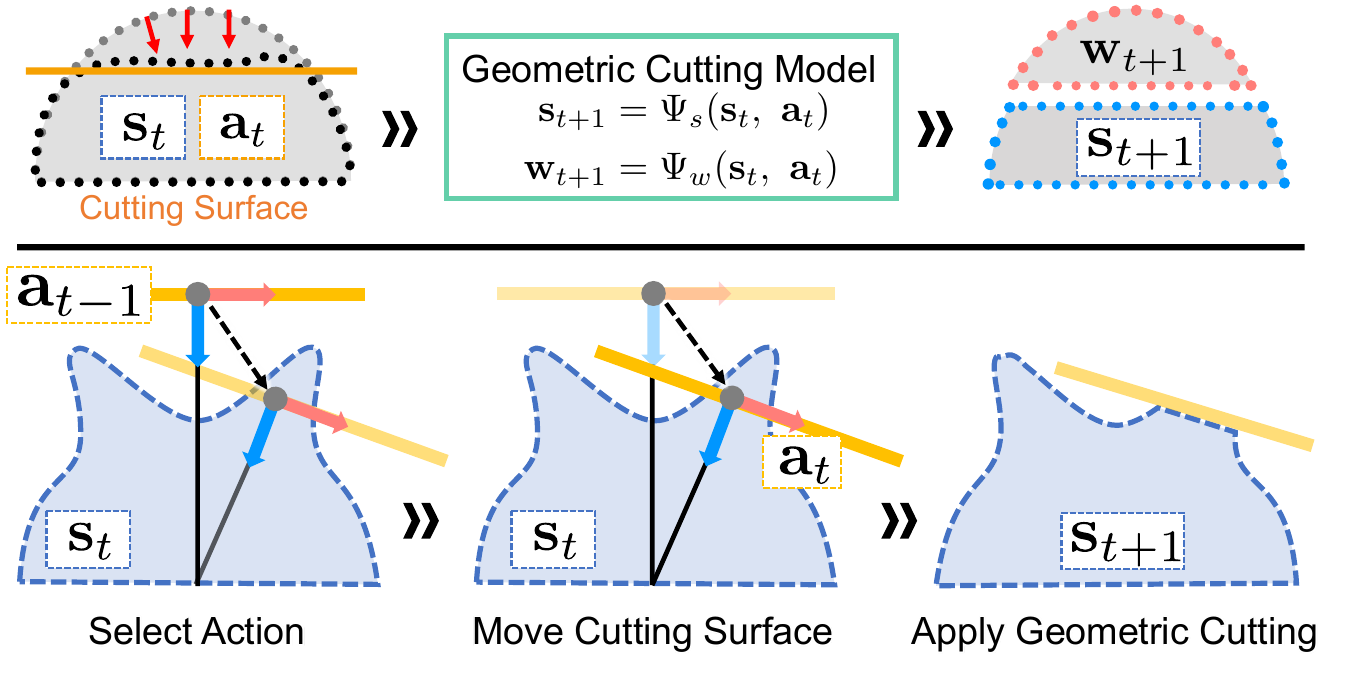}
    \caption{Action space for grinding with a small removal volume.
    \textbf{Top}: Shape transition by cutting surface. 
    \textbf{Bottom}: Transition examples of cutting surface.}
    \label{fig:cutting_surface_move}
    \vspace{-4truemm}
\end{figure}
\subsection{Cutting Sequence Diffuser Framework}\label{subsection:CSD_framework}
The proposed method consists of the following three major phases (\figurename\ref{fig:block_diagram}): 1) Data Collection: 
We collect \(K\) episodes of random state action sequences constrained by a small removal volume. Data collection is conducted by a simulator that implements a geometric cutting model. The details of the action constraints are described in \S \ref{subsection:action_constrain}.
2) Cutting Sequence Diffuser Training: A diffusion model is trained to generate trajectories with planning horizon \(H\) using the collected data. 
During the training, high-dimensional shapes are reduced by a variational auto-encoder, as described in \S \ref{subsection:shape_encoding}. 
3) Deployment:  
The planned cutting sequences are integrated with closed-loop control to ensure robust action execution, similar to MPC. 
CSD plans \(H\)-horizon cutting sequences based on the current observed shape, the cost function, and the two-step guidance described in \S \ref{subsection:diffusion_cost_design}.
Then, \(M\)-step actions \((1 \leq M \leq H)\) from the planned cutting sequence are executed as control inputs for the robot.
By repeating this process until task horizon \(T\), 
object-shaping can be achieved by grinding with a diffusion model.

\subsection{Sim-to-real Transferable Data Collection} \label{subsection:action_constrain}
A geometric cutting model is only applicable when the removal volume in one step is sufficiently small.
Therefore, random state action sequences for diffusion model learning are collected under the following constraints:
\begin{subequations} \label{eq:ctrl_constrain}
\begin{align}
& \text{sample}
& &{\mathbf s}_{t+1}^{}= {\Psi_s}(\mathbf{s}_t,~{\mathbf{a}}_t), \nonumber \\
& \text{subject to} 
& & f_{\mathrm{col}}({\mathbf s}_{t},\mathbf{a}_t) \leq \epsilon_\text{vol}, \label{eq:ctrl_constrain:g1} \\
&&& f_{\mathrm{col}}({\mathbf s}_{\mathrm{ref}},\mathbf{a}_t) \leq d_\text{vol}.\label{eq:ctrl_constrain:g2}
\end{align}
\end{subequations}
Where \(f_{\mathrm{col}}(\cdot)\) returns 0 if no removal volume exists; otherwise it returns the removal volume.
\(\epsilon_{\text{vol}}\) is the upper bound of the removal volume at one step.
\equationname\ref{eq:ctrl_constrain:g2} constrains the cutting action that overcuts reference shape \(\mathbf{s}_{\mathrm{ref}}\) to improve the efficiency of data collection.
\(d_\mathrm{vol}\) is the volume threshold that overcuts the reference shape.
\begin{figure}[t!]
    \centering
    \includegraphics[clip,width=1.0\linewidth]{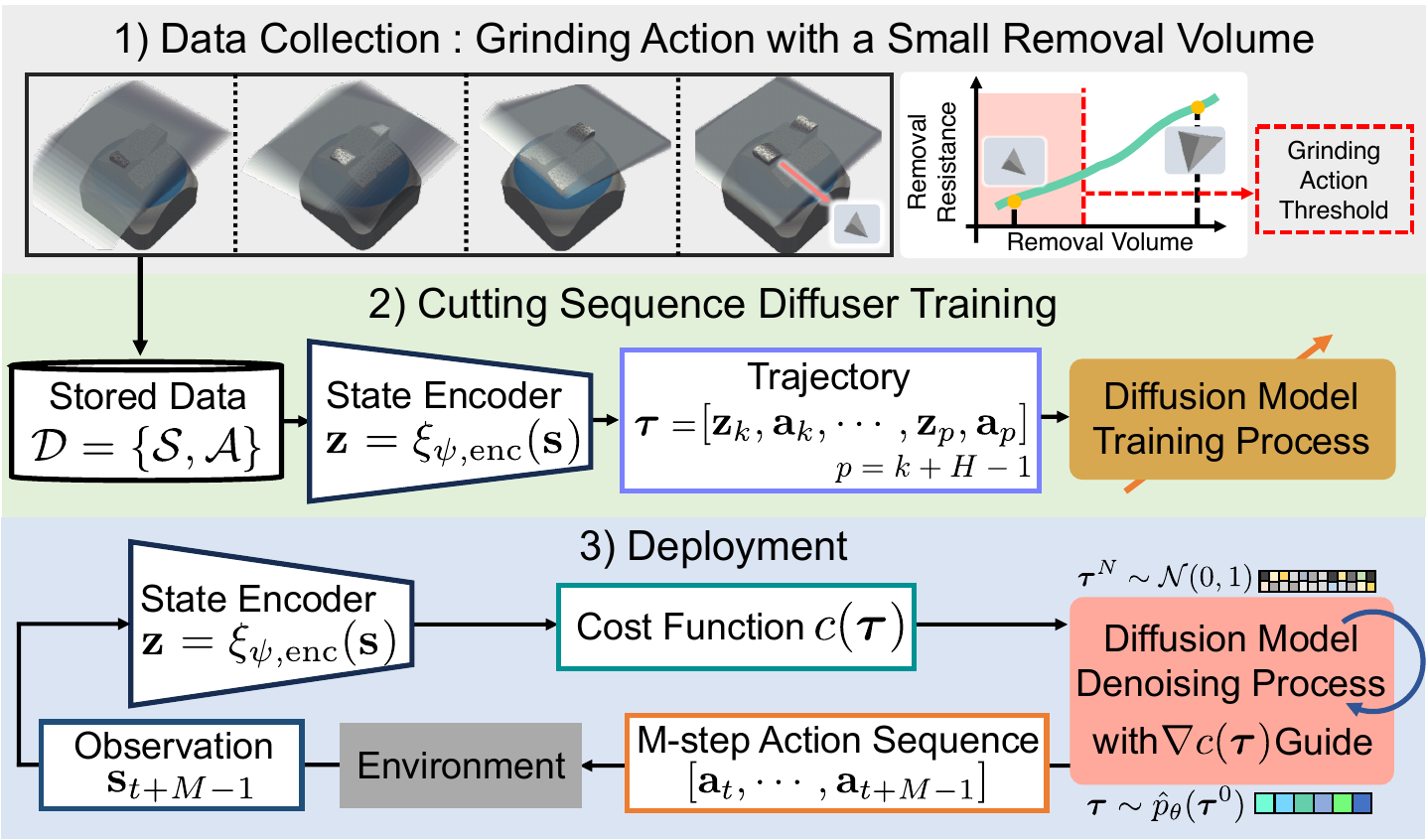}
    \caption{Framework of Cutting Sequence Diffuser.
    A diffusion model is trained by collected data under grinding action with a small removal volume. 
    At the deployment, planned $M$-step action (cutting) sequences are executed based on the current observed shape.}
    \label{fig:block_diagram}
    \vspace{-4truemm}
\end{figure}
\subsection{Latent Encoding of Shape States}\label{subsection:shape_encoding}
In grinding processes, shape representations such as point clouds and depth images used as states are high-dimensional.
Directly using such high-dimensional states increases the training costs of a diffusion model. 
Therefore, a Variational Auto-Encoder (VAE) \(\xi_{\psi,\mathrm{enc}}\) with parameters \(\psi\) is used to compress state \(\mathbf{s}_t\) into a latent feature \((\mathbf{z}_t = \xi_{\psi,\mathrm{enc}}(\mathbf{s}_t), \mathbf{z}_t \in \mathbb{R}^{dz})\).
Where \({dz}\) is the dimension of the latent features. 
Thus, trajectory \(\boldsymbol{\tau}\) used in a diffusion model is defined as \(\boldsymbol{\tau} \coloneqq [\mathbf{z}_{t}, \mathbf{a}_{t}, \ldots, \mathbf{z}_{t+H-1}, \mathbf{a}_{t+H-1}]\).
In our experiments shown below, we employed point cloud \(\mathcal{U}_t:=[{\mathbf u}_{m,t}]_{m=1}^{D}\subseteq \mathbb{R}^{3}\) as the state, where \(D\) is the number of points and \(\mathbf{u}_{m,t}\) is the \(m\)-th particle position.

\subsection{Guided Sampling for Cutting Sequence Generation }\label{subsection:diffusion_cost_design}
\subsubsection{Cost Design}
This section describes the types of cost functions used as guides for trajectory generation by a diffusion model.
Total cost \(c(\boldsymbol{\tau})\) for \(t\) to \(t+H-1\) is calculated by summing each cost: \(c(\boldsymbol{\tau})= \lambda_{\mathrm{sm}}c_{\mathrm{sm}}(\boldsymbol{\tau})+\sum_{l=t}^{t+H-1}\sum_{k}\lambda_{k}c_k(\boldsymbol{\tau}[l])\), \(k=\{\mathrm{state},\mathrm{col},\mathrm{vol}\}\).
Where \(\lambda\) are the weights of each cost.

\textbf{State constrain cost:}
This cost generates a trajectory that satisfies the state constraint \(\mathbf{c}_{l}\), such as current shape \(\mathbf{s}_\mathrm{current}\) and target shape \(\mathbf{s}_\mathrm{target}\).
Observed state distributions in diffusion models are represented by a Dirac delta function. Hence, state constraint cost \(c_\mathrm{state}\) is expressed as follows:
\begin{equation}
c_\mathrm{state}(\boldsymbol{\tau}[l])=\delta_{\mathbf{c}_{l}}(\mathbf{s}_{l})=
  \begin{cases}
    -\infty &\text{if}~\mathbf{c}_{l}=\mathbf{s}_{l}, \\
    0       &\text{otherwise}.
  \end{cases}
\end{equation}
In the implementation, if \(\mathbf{s}_l\) satisfies the state constraint, we directly replace \(\mathbf{s}_l\) with a state constraint value \(\mathbf{c}_l\) (e.g., \(\mathbf{s}_\mathrm{current}\) or \(\mathbf{s}_\mathrm{target}\)) in all the denoising processes, as in \cite{janner2022diffuser}.

\textbf{Over-cutting cost:}
Since the grinding process is irreversible, over-cutting target shape \(\mathbf{s}_\mathrm{target}\) is undesirable.
Therefore, we introduce over-cutting cost \(c_\mathrm{col}\):
\begin{align}
&c_\mathrm{col}(\boldsymbol{\tau}[l])=\begin{cases}
    f_{\mathrm{col}}({\mathbf s}_{\mathrm{target}},\mathbf{a}_l)&\text{if}~f_{\mathrm{col}}({\mathbf s}_{\mathrm{target}},\mathbf{a}_l)>0,\\
    0&\text{otherwise}.
  \end{cases}
\end{align}

\textbf{Cut volume limit cost:}
This cost constrains the removal volume at one step.
If removal volume exceeds \(\delta_{\mathrm{vol}}\), the following cost $c_\mathrm{vol}$ is imposed:
\begin{align}
&c_\mathrm{vol}(\boldsymbol{\tau}[l])= \begin{cases}
     f_{\mathrm{col}}({\mathbf s}_{l},\mathbf{a}_l) -\delta_{\text{vol}} &\text{if}~f_{\mathrm{col}}({\mathbf s}_{l},\mathbf{a}_l) \geq \delta_{\text{vol}},\\
    0&\text{otherwise}.
  \end{cases}
\end{align}

\textbf{Action smoothness cost:}
To suppress the vibrations of the robot's end-effector and promote the planning of smooth cutting sequences, we constrain the travel length of the cutting surface as follows:
\begin{gather}
c_{\text{sm}}(\boldsymbol{\tau}) = \sum_{l=t}^{t+H-2}\sum_{j}
d_{\text{len}}({a}_{l}^{j},{a}_{l+1}^{j}), \nonumber \\
d_{\text{len}}({a}_{l}^{j},{a}_{l+1}^{j})=
\begin{cases} 
\| {a}_{l}^{{j}} - {a}_{l+1}^{j} \|_2^2 &\hspace{-1.9truemm} \text{if}~| {a}_{l}^{{j}} - {a}_{j+1}^{j}|\hspace{-.4truemm} > \hspace{-.4truemm} a_{}^{j_{\text{max}}}, \\
0 &\hspace{-2.0truemm}\text{otherwise}.
\end{cases}
\label{eq:trajectory_smoothness_cost}
\end{gather}
Where \(a_l^j\) is the \(j\)-th action value at time step \(l\) and \({a}_{}^{j_{\text{max}}}\) is the upper bound of the travel length for each action dimension. The variable \(j\) indicates the action index.

In the general grinding process, overcutting the target shape and exceeding the volume limit that can be removed in one step is fatal. Therefore, it would be preferable to set the weights to ensure that the \(c_{\mathrm{col}}\) and \(c_{\mathrm{vol}}\) costs are dominant.

\subsubsection{Trajectory Generation with Two-step Guide}\label{subsubsection:double_guidance}
The proposed method increases task horizon \(T\) by constraining the removal volume to reduce the reality gap.
Therefore, as shown in \S \ref{subsection:CSD_framework}, the task horizon is divided into practical planning horizon lengths \(H\).
However, the grinding process is a goal-conditioned trajectory planning problem where only current shape \(\mathbf{s}_{t}\) and final target shape \(\mathbf{s}_{\mathrm{target}}\) can be used as state constraints.
Thus, if we naively set the final target shape as the end of planning horizon constraint \(t+H-1\),
CSD tries to achieve object shaping, which normally requires \( T \) steps, within a planning horizon \( H \). 
This typically occurs when the planning horizon is shorter than the task horizon (i.e., \( t + H - 1 \ll T \)).
Hence, such an approach can generate cutting sequences that remove a large volume at once, potentially causing significant removal resistance.

To address this issue, we execute trajectory planning with two-step guidance.
Initially, at \(t=0\), we set planning horizon \(H\) equal to task horizon \(T\) and generate a trajectory for the entire task horizon to obtain the reference shape states for each time step. 
Next, we plan the \(H\)-horizon trajectory by referring to the obtained shape as the state constraint at the end of the planning horizon, \(t+H-1\). Thus, we use the shape state obtained by planning the trajectory with the entire task horizon as the state constraint at the end of the planning horizon. 
This approach allows for the suppression of cutting actions with large removal volumes at once. 
As a result, it may reduce removal resistance and promote smooth trajectory generation.

\section{Robotic Rough-Grinding Environments}
Figure~\ref{fig:eye_cath} shows our robotic rough-grinding system.
A grinding belt is fixed in the environment, and the object is mounted on the 6-DoF robot's end-effector. 
The robot moves to the front of the 3D vision sensor and captures a point cloud as the object's shape.
Robot action \({\mathbf{a}_t}\) manipulates the coordinates of the object around the work origin, which is an arbitrary distance from the grinding belt.
It is equivalent to specifying the cutting surface of the object.
The 2-DoF of the parallel translation and the 1-DoF rotation around a perpendicular axis can be ignored because the grinding belt is sufficiently wide.
As a result, the robot's action becomes \(\mathbf{a}_t:=[\rm{roll}_t,\rm{pitch}_t,\rm{z}_t]\in{{\mathbb R}}^{j=3}\).

Figure~\ref{fig:real_env} shows the initial and target shapes used for the experiments, all of which were fabricated by a 3D printer. 
The gray part is the base, the blue part is the target shape, and the white part is the shape to be removed by grinding.
We prepared white parts printed with Acrylonitrile Styrene Acrylate (ASA) filament and PolyCarbonate (PC) filament, the latter of which is a harder material than ASA.

\begin{figure}[t!]
    \centering
    \includegraphics[clip,width=1.0\linewidth]{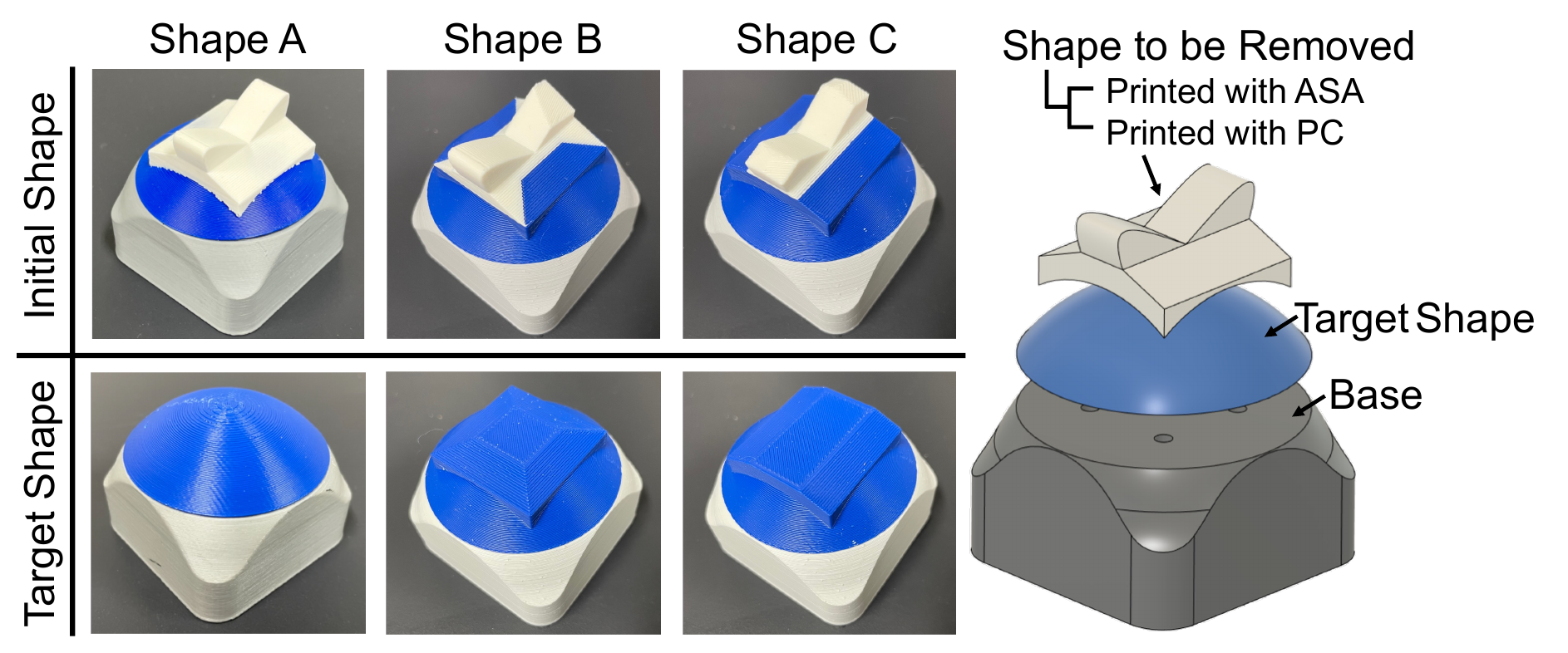}
    \caption{
    Configurations of initial and target shapes for experiments.
    }
    \label{fig:real_env}
    \vspace{-5truemm}
\end{figure}
\begin{figure*}[t!]
\centering
\begin{minipage}[t]{1.96\columnwidth}
    \centering
        \includegraphics[clip,width=\columnwidth]{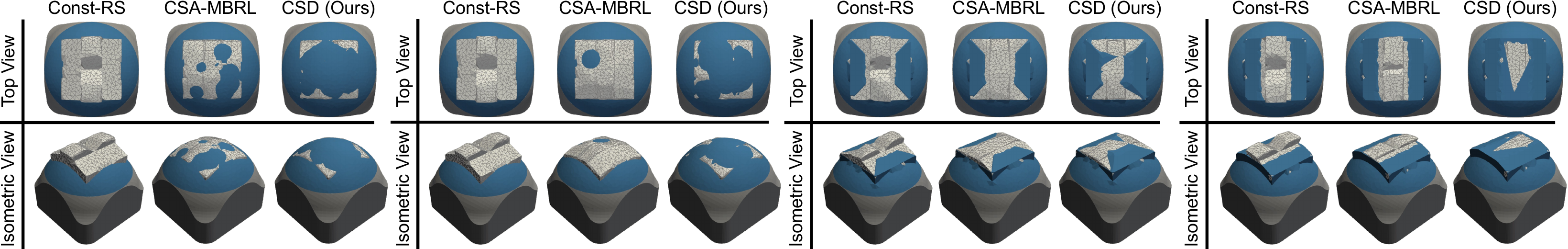}
    \begin{minipage}[b]{0.24\columnwidth}
        \centering
        \subcaption{Shape A with ASA}
    \end{minipage}
    \begin{minipage}[b]{0.24\columnwidth}
        \centering
        \subcaption{Shape A with PC}
    \end{minipage}
    \begin{minipage}[b]{0.24\columnwidth}
        \centering
        \subcaption{Shape B with ASA}
    \end{minipage}
    \begin{minipage}[b]{0.24\columnwidth}
        \centering
        \subcaption{Shape C with ASA}
    \end{minipage}
    \caption{
    Rendered shape after grinding for each method in simulation experiments.}
    \label{fig:sim_result:4_result_concat_sim}
\end{minipage}
\vspace{-2.0truemm}
\end{figure*}
\begin{table*}[t!]
\caption{Comparison of shape error and action planning time{\,[s]} for each method in simulation experiments.}
\label{tab:sim_task_perfomance_comp}
\centering
\setlength{\tabcolsep}{3.7pt}
\scalebox{0.93}{
\begin{tabular}{llc|cc|cc|cc}
\toprule
&\multicolumn{2}{c}{\makecell{Shape A with ASA}}&\multicolumn{2}{c}{\makecell{Shape A with PC}}&\multicolumn{2}{c}{\makecell{Shape B with ASA}}&\multicolumn{2}{c}{\makecell{Shape C with ASA}} \\
\cmidrule(lr){2-3}
\cmidrule(lr){4-5}
\cmidrule(lr){6-7}
\cmidrule(lr){8-9}
Method & Shape Error & Planning Time{\,[s]} & Shape Error & Planning Time{\,[s]}& Shape Error & Planning Time{\,[s]}& Shape Error& Planning Time{\,[s]}\\
\midrule
Const-RS  & 1.42 {$\pm 0.004$} & 691.82 {$\pm 9.03$} & 1.42 {$\pm 0.002$} & 691.73 {$\pm 8.61$} 
          & 1.89 {$\pm 0.01$} & 435.41 {$\pm 1.95$} & 1.60 {$\pm 0.004$} & 289.75 {$\pm 5.04$} \\
CSA-MBRL  & 0.96 {$\pm 0.01$} & 140.21 {$\pm 1.16$} & 1.07 {$\pm 0.02$} & 140.94 {$\pm 0.98$}
          & 1.52 {$\pm 0.05$} & 93.89 {$\pm 0.88$}  & 1.40 {$\pm 0.03$} & 70.87 {$\pm 0.94$}  \\
CSD (Ours) & \textbf{0.94} {$\pm 0.02$} & \textbf{46.98} {$\pm 0.81$}  & \textbf{0.95} {$\pm 0.01$} & \textbf{47.55} {$\pm 0.91$}
           & \textbf{1.31} {$\pm 0.02$} & \textbf{39.39} {$\pm 0.81$} & \textbf{1.18} {$\pm 0.02$} & \textbf{28.17} {$\pm 0.64$} \\
\bottomrule
\end{tabular}
}
\vspace{-2truemm}
\end{table*}
\begin{table*}[t]
\centering
\caption{Comparison of generated trajectory by diffusion model with/without a guide function in simulation experiments.}
\label{tab:sim_guide_abration}
\setlength{\tabcolsep}{2.5pt}
\scalebox{.93}{
\begin{tabular}{lcccc|cccc}
\toprule
&\multicolumn{4}{c}{\makecell{Shape A with ASA}}&\multicolumn{4}{c}{\makecell{Shape A with PC}}\\
\cmidrule(lr){2-5}
\cmidrule(lr){6-9}
Method & Shape Error & Over-cutting & Action Smoothness & Cut Volume Limit & Shape Error & Over-cutting & Action Smoothness & Cut Volume Limit\\
\midrule
CSD w/o Guide & 0.97 {$\pm 0.03$} & 16.17 {$\pm 4.88$} & 1.52 {$\pm 0.04$} & 0.09 {$\pm 0.01$}
                               & 0.98 {$\pm 0.05$} & 20.17 {$\pm 7.15$} & 1.51 {$\pm 0.10$} & 0.10 {$\pm 0.02$} \\
CSD w/ Guide   & \textbf{0.94} {$\pm 0.02$} & \textbf{5.67} {$\pm 2.29$} & \textbf{0.76} {$\pm 0.07$} & \textbf{0.05} {$\pm 0.01$}
                             & \textbf{0.95} {$\pm 0.01$} & \textbf{3.83} {$\pm 1.34$} & \textbf{0.64} {$\pm 0.04$} & \textbf{0.04} {$\pm 0.01$} \\
\bottomrule
\end{tabular}
}
\vspace{-4truemm}
\end{table*}
\begin{table}[t]
\centering
\caption{ Effects of trajectory generation with two-step guide in simulation
experiments.}
\label{tab:sim_2_step_guide_compariosn}
\setlength{\tabcolsep}{3.7pt}
\scalebox{.93}{
\begin{tabular}{lc|c}
\toprule
&\multicolumn{2}{c}{\makecell{Cut Volume Limit}} \\
\cmidrule(lr){2-3}
Method & Shape A with ASA & Shape A with PC \\
\midrule
CSD w/ One-step Guide & 0.074 {$\pm 0.006$}& 0.077 {$\pm 0.011$}\\
CSD w/ Two-step Guide & 0.051 {$\pm 0.006$}& 0.044 {$\pm 0.005$}\\
\bottomrule
\vspace{-5truemm}
\end{tabular}
}
\end{table}
\begin{figure}[t!]
    \centering
    \includegraphics[clip,width=0.95\linewidth]{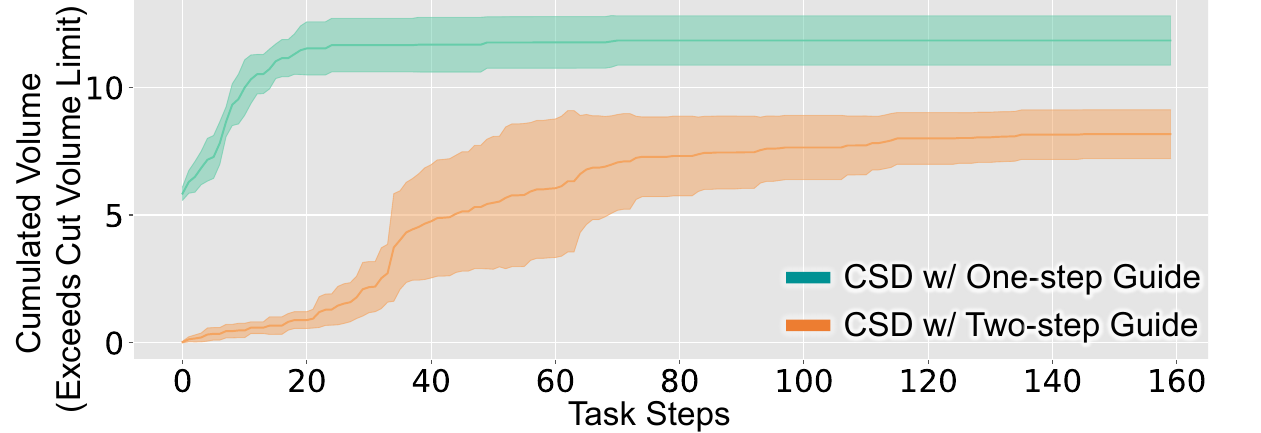}
    \caption{Comparison of cumulated volume which exceeds cut volume limit when processing Shape A with ASA in simulation experiments.}
    \label{fig:sim_result:double_step_guide_comp_sim}
    \vspace{-2truemm}
\end{figure}
\begin{table}[t!]
    \caption{Comparison of action planning methods within the proposed method in simulation experiments.}
\label{tab:sim_cvae_task_perfomance_comp}
\centering
\setlength{\tabcolsep}{3.7pt}
\scalebox{1.0}{
\begin{tabular}{llc}
\toprule
&\multicolumn{2}{c}{\makecell{Shape A with ASA}} \\
\cmidrule(lr){2-3}
Planning Method & Shape Error & Planning Time{\,[s]} \\
\midrule
CVAE             & 1.08 {$\pm 0.05$} & \textbf{23.37} {$\pm 0.50$}\\
Diffusion & \textbf{0.94} {$\pm 0.02$} & {46.98} {$\pm 0.81$}\\
\bottomrule
\end{tabular}}
\vspace{-6truemm}
\end{table}
\section{SIMULATION EXPERIMENT}
In our simulation experiment, we evaluated the proposed method with an environment that implemented a virtual grinding resistance as a proxy for the real world.
This experiment had four objectives. First, we compared the shape error after the processing and the action planning time to the baseline methods. Second, we compared the action planner used in the proposed method. Third, we investigated its applicability for environments with different grinding resistances and different initial and target shapes. Fourth, we evaluated the effectiveness of the guided trajectory generation with a designed cost function. 
Additional experiments and parameter settings are available on our project page: https://t-hachimine.github.io/csd/.

\subsection{Simulation Environments} 
We used Open3D \cite{Zhou2018} and Pyvista \cite{sullivan2019pyvista} for the rendering.
The initial and target shapes were created by CAD and converted to the point cloud.
We implemented a virtual grinding resistance model, depending on the robot's action and the removal volume.
In the simulation experiment, the shape deformation due to the grinding resistance was reproduced as the deviation of the cutting surface.
We prepared two virtual grinding resistance models (ASA and PC) to reproduce different grinding resistances with the material. The removal volume was calculated by the sum of the Euclidean distance between the removal shape and the cutting surface divided by the number of removal shape point clouds.

\subsection{Baseline Methods}\label{subsub:comparison_method}
\textbf{Const-RS}: This method samples action with a small removal volume similar to the proposed method and executes action planning by MPC. 
We used a random shooting method \cite{nagabandi2018neural} to optimize the cost function of MPC.
\textbf{CSA-MBRL}: The MBRL method learns a shape transition model by considering the grinding resistance using real data and executes action planning by MPC \cite{10214100}.
\textbf{CVAE}: This method uses Conditional VAE (CVAE) \cite{carvalho2023motion,ichter2018learning} as an action planner instead of a diffusion model.
The encoder and decoder networks used the same structure as a diffusion model.
During training, the conditions were set as the start and end states of the planning horizon. 
At the planning time, trajectories were conditioned by the current and target states. 
Finally, action sequences were derived by optimizing the likelihood against other costs using the gradient descent method. 

\subsection{Data Collection and Action Planning Settings}
We randomly sampled actions with \(\epsilon_{\mathrm{vol}}=1.0\) and did not overcut the target shape of Shape A as a reference shape \({\mathbf s}_{\mathrm{ref}}\) by more than \(d_{\mathrm{vol}}=0.3\). We prepared 9K trajectories with 251 steps per episode.
Shape state ${\mathbf{s}_t}$ was compressed into latent features $(dz=64)$ by a VAE with a PointNet structure \cite{qi2017pointnet}. 
Following the same learning procedure as in \cite{carvalho2023motion,janner2022diffuser}, we calculated the training loss of a 1D-Unet diffusion model using \(c_\mathrm{state}\).
Finally, we trained it in 2M steps with batch size $128$, $H=32$, and diffusion step $i=64$.

\begin{figure*}[t!]
\centering
\begin{minipage}[t]{1.96\columnwidth}
    \centering
        \includegraphics[clip,width=\columnwidth]{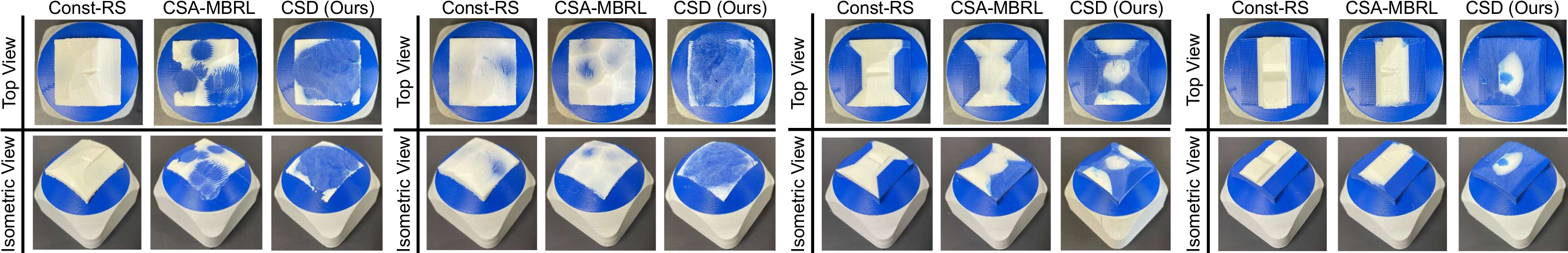}
    \begin{minipage}[b]{0.24\columnwidth}
        \centering
        \subcaption{Shape A with ASA}
    \end{minipage}
    \begin{minipage}[b]{0.24\columnwidth}
        \centering
        \subcaption{Shape A with PC}
    \end{minipage}
    \begin{minipage}[b]{0.24\columnwidth}
        \centering
        \subcaption{Shape B with ASA}
    \end{minipage}
    \begin{minipage}[b]{0.24\columnwidth}
        \centering
        \subcaption{Shape C with ASA}
    \end{minipage}
    \caption{
    Captured shape after grinding for each method in robot experiments.}
    \label{fig:real_result:4_result_concat}
\end{minipage}
\vspace{-2.0truemm}
\end{figure*}
\begin{table*}[t!]
\caption{Comparison of shape error and action planning time{\,[s]} for each method in robot experiments.}
\label{tab:real_task_perfomance_comp}
\centering
\setlength{\tabcolsep}{3.7pt}
\scalebox{.95}{
\begin{tabular}{llc|cc|cc|cc}
\toprule
&\multicolumn{2}{c}{\makecell{Shape A with ASA}}&\multicolumn{2}{c}{\makecell{Shape A with PC}}&\multicolumn{2}{c}{\makecell{Shape B with ASA}}&\multicolumn{2}{c}{\makecell{Shape C with ASA}} \\
\cmidrule(lr){2-3}
\cmidrule(lr){4-5}
\cmidrule(lr){6-7}
\cmidrule(lr){8-9}
Method & Shape Error & Planning Time{\,[s]} & Shape Error & Planning Time{\,[s]}& Shape Error & Planning Time{\,[s]}& Shape Error& Planning Time{\,[s]}\\
\midrule
Const-RS     & 3.05 {$\pm 0.15$} & 1087.00 {$\pm 4.20$} & 2.58 {$\pm 0.26$} & 1090.55 {$\pm 10.82$} & 1.90 {$\pm 0.19$} & 900.00 {$\pm 15.98$} & 2.00 {$\pm 0.10$} & 662.28 {$\pm 6.32$} \\
CSA-MBRL     & 1.40 {$\pm 0.09$} & 89.13 {$\pm 2.41$} & 1.68 {$\pm 0.16$} & 87.48 {$\pm 2.04$} & 1.54 {$\pm 0.02$} & 90.33 {$\pm 1.46$} & 1.52 {$\pm 0.20$} & 68.02 {$\pm 0.96$} \\
CSD (Ours)   & \textbf{1.28} {$\pm 0.06$} & \textbf{49.61} {$\pm 1.35$} & \textbf{1.27} {$\pm 0.03$} & \textbf{45.87} {$\pm 1.76$} & \textbf{1.21} {$\pm 0.02$} & \textbf{35.46} {$\pm 0.35$} & \textbf{1.16} {$\pm 0.02$} & \textbf{28.63} {$\pm 0.61$} \\
\bottomrule
\end{tabular}
}
\vspace{-2truemm}
\end{table*}
\begin{table*}[t]
\centering
\caption{Comparison of generated trajectory by diffusion model with/without a guide function in robot experiments.}
\label{tab:real_guide_abration}
\setlength{\tabcolsep}{2.6pt}
\scalebox{.93}{
\begin{tabular}{lcccc|cccc}
\toprule
&\multicolumn{4}{c}{\makecell{Shape A with ASA}}&\multicolumn{4}{c}{\makecell{Shape A with PC}}  \\
\cmidrule(lr){2-5}
\cmidrule(lr){6-9}
Method & Shape Error & Over-cutting & Action Smoothness & Cut Volume Limit & Shape Error & Over-cutting & Action Smoothness & Cut Volume Limit\\
\midrule
CSD w/o Guide & 1.52 {$\pm 0.20$} & 20.00 {$\pm 4.97$} & 1.42 {$\pm 0.15$} & 0.11 {$\pm 0.01$} & 2.46 {$\pm 0.63$} & 26.00 {$\pm 0.82$} & 1.31 {$\pm 0.06$} & 0.12 {$\pm 0.01$} \\
CSD w/ Guide & \textbf{1.28} {$\pm 0.06$} & \textbf{4.00} {$\pm 1.63$} & \textbf{0.62} {$\pm 0.01$} & \textbf{0.05} {$\pm 0.01$} & \textbf{1.27} {$\pm 0.03$} & \textbf{4.00} {$\pm 1.63$} & \textbf{0.73} {$\pm 0.01$} & \textbf{0.05} {$\pm 0.01$} \\
\bottomrule
\vspace{-8truemm}
\end{tabular}
}
\end{table*}
\begin{table}[t]
\centering
\caption{Effects of trajectory generation with two-step guide in robot experiments.}
\label{tab:real_2_step_guide_compariosn}
\setlength{\tabcolsep}{3.7pt}
\scalebox{.93}{
\begin{tabular}{lc|c}
\toprule
&\multicolumn{2}{c}{\makecell{Cut Volume Limit}} \\
\cmidrule(lr){2-3}
Method & Shape A with ASA & Shape A with PC\\
\midrule
CSD w/ One-step Guide & 0.094 {$\pm 0.012$}& 0.086 {$\pm 0.005$}\\
CSD w/ Two-step Guide & 0.046 {$\pm 0.008$}& 0.048 {$\pm 0.006$}\\
\bottomrule
\vspace{-4truemm}
\end{tabular}
}
\end{table}
\begin{figure}[t!]
    \centering
    \includegraphics[clip,width=1.0\linewidth]{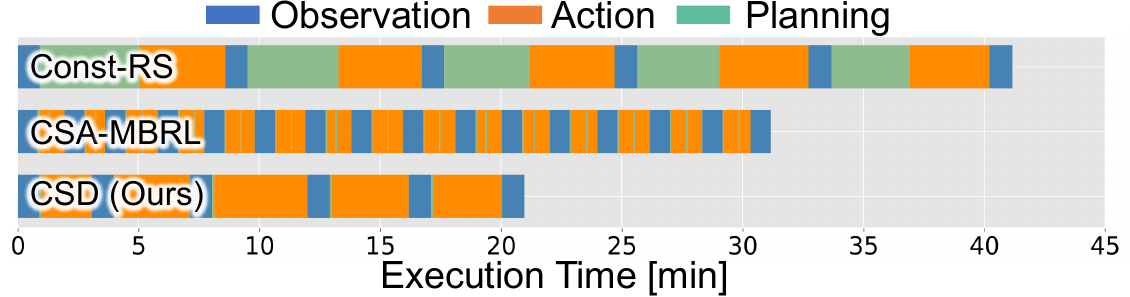}
    \caption{Task execution time comparison of each method when processing Shape A with ASA.}
    \label{fig:real_result:env_1_object_A_timeshart}
    \vspace{-6truemm}
\end{figure}
\subsection{Simulation Results}
\subsubsection{Evaluation with Baseline Methods}
Figure~\ref{fig:sim_result:4_result_concat_sim}~(a) shows an example of the shape after processing Shape A with ASA. 
\tablenameref\ref{tab:sim_task_perfomance_comp} compares the shape error and the action planning time for six trials of each method.  
In the following experiment, we used Chamfer discrepancy  \cite{Nguyen2021PointsetDF} for the shape error calculation:
\begin{equation}
d_{\rm CD}(\mathbf{s}_1,\mathbf{s}_2)=
\frac{1}{|\mathbf{s}_1|}
\sum_{\mathbf{x} \in \mathbf{s}_1}\underset {\mathbf{y} \in \mathbf{s}_2}{\rm min} \|\mathbf{x}-\mathbf{y}\|_{2}^{2}+\frac{1}{|\mathbf{s}_2|}
\sum_{\mathbf{y} \in \mathbf{s}_2}\underset {\mathbf{x} \in \mathbf{s}_1}{\rm min} \|\mathbf{x}-\mathbf{y}\|_{2}^{2},
\label{eq:champer_distances}
\end{equation}
where \(\mathbf{x}\) and \(\mathbf{y}\) are the positions of the individual points in the point clouds.
This metric evaluates the error using the Euclidean distance between the nearest neighbor points of two point clouds. It indicates that the shape error is rarely zero unless the two point clouds are perfectly aligned.
\textbf{Const-RS} and \textbf{CSD} were performed with $T=160$ task steps, a planning horizon of $H=32$, and $M=32$ control inputs.
\textbf{CSA-MBRL} was performed with $T=40$, $H=5$, and $M=1$.
\textbf{CSA-MBRL} deﬁnes a single-step action as follows: 1) Transitioning from the work origin to the target cutting surface, 2) Holding the target cutting surface for a speciﬁc duration, and 3) Returning to the work origin.
This definition allows each step to cover a larger range of tasks, leading to shorter \(T\), \(H\), and \(M\). 
Additionally, \(T\), \(H\), and \(M\) were set based on \cite{10214100} to maintain consistency with prior experiments and ensure comparability.
These results show that our proposed method can quickly plan long-horizon action sequences and achieve comparable shape error as \textbf{CSA-MBRL}, which learns the shape transition model from real data.

\subsubsection{Comparison of Action Planning Methods within the Proposed Method}
\tablenameref\ref{tab:sim_cvae_task_perfomance_comp} compares the action planner within the proposed method.
The former with \textbf{CVAE} had a shorter planning time than the diffusion model. However, it had larger shape error, indicating the suitability of employing a diffusion model for the proposed method, where grinding accuracy is important even if action planning takes some time.

\subsubsection{Applicability of Proposed Method}
Figures~\ref{fig:sim_result:4_result_concat_sim}~(b)-(d) and \tablenameref\ref{tab:sim_task_perfomance_comp} show the results of grinding Shape A with PC and Shapes B and C with ASA.
These results indicate that the proposed method can be applied to environments with different grinding resistances and different initial and target shapes by combining actions with a small removal volume through flexible trajectory generation using a diffusion model.

\subsubsection{Effects of Guided Diffusion}
\tablenameref\ref{tab:sim_guide_abration} compares the task performances with/without a guide.
Four evaluation metrics were used for the task performance.
Over-cutting is the number of times the target shape was excessively cut.
The cut volume limit is obtained by dividing the cumulative sum of the volumes that exceed the removal volume threshold \(\delta_{\mathrm{vol}}\) by the number of task steps.
Refer to \eqrefs{eq:trajectory_smoothness_cost}{eq:champer_distances} concerning shape error and action smoothness.
These results show that a guided trajectory generation with a cost function improves each evaluation metric.

\tablenameref\ref{tab:sim_2_step_guide_compariosn} compares the cut volume limit with/without the two-step guide. \figurename\ref{fig:sim_result:double_step_guide_comp_sim} shows the transition of the cumulative sum of the volumes that exceeded the cut volume limit.
In the case of a one-step guide, planning actions exceeded the cut volume limits around the initial step. However, the two-step guide suppressed the excesses of the cut volume limits.

\section{REAL ROBOT EXPERIMENT}
We evaluated the proposed method in a real robot environment because the grinding resistance was basically proportional to the removed volume, however, in the actual grinding process, it was also generated from other effects (e.g., friction heat).
In the robot experiment, cutting sequences were generated by a diffusion model trained with the same settings as the simulation experiment.
The robot followed each linearly interpolated action (cutting surface) at a constant velocity by position control.
The shape coordinate error between the simulation and the real world was calibrated in advance.

\subsection{Results}
\subsubsection{Evaluation with Baseline Methods}
Figure~\ref{fig:real_result:4_result_concat}~(a) shows an example of the shape after processing Shape A with ASA. 
\tablenameref\ref{tab:real_task_perfomance_comp} compares the shape error and the action planning time for three trials of each method. 
\textbf{Const-RS} and \textbf{CSD} were performed with $T=160$ task steps, a planning horizon of $H=32$, and $M=32$ control inputs.
\textbf{CSA-MBRL} was performed with $T=30$, $H=5$, and $M=1$.
\textbf{CSA-MBRL} observed the shape every two steps to reduce the task execution time.
The shape observation in the real robot experiments required the robot to be moved in front of the 3D vision sensor (\figurename\ref{fig:eye_cath}) and to take multiple measurements to obtain an accurate shape, which was time-consuming.
Therefore, the predicted shape (one step ahead) was used as the observed shape where no shape observation was performed, and the action was planned again.
These results show that, similar to the simulation results, the proposed method can quickly plan long-horizon action sequences and achieve comparable shape error as \textbf{CSA-MBRL}, which learns the shape transition model from real data.

Figure~\ref{fig:real_result:env_1_object_A_timeshart} compares the observations, action planning, and action execution times required to process Shape A with ASA.
\textbf{Const-RS} required longer time to optimize a long-horizon action.
\textbf{CSA-MBRL}, required frequent shape observations to reduce the shape error.
In contrast, the proposed method required less task execution time due to its ability to quickly plan long-horizon action sequences.

\subsubsection{Applicability of Proposed Method}
Figures~\ref{fig:real_result:4_result_concat}~(b)-(d) and \tablenameref\ref{tab:real_task_perfomance_comp} show the results of grinding Shape A with PC and Shapes B and C with ASA.
The proposed method can be applied to different materials and different initial and target shapes by combining actions with a small removal volume through flexible trajectory generation using a diffusion model.

\subsubsection{Effects of Guided Diffusion}
\tablenameref\ref{tab:real_guide_abration} compares the task performances with/without a guide.
These results show that a guided trajectory generation with a cost function improves each evaluation metric.
Additionally, \tablenameref\ref{tab:real_2_step_guide_compariosn} compares the cut volume limit with/without the two-step guide. 
We confirmed that it suppressed the excesses of the cut volume limits.

\section{DISCUSSION}
The evaluation of this study focused on applicability to different materials and initial and target shapes. Therefore, investigating the applicability of different belt rotation speeds and abrasive grain types is interesting future work.
Since the proposed method reduced removal resistance by constraining the removal volume, 
it may be adaptable by appropriately setting the cut volume limit cost used for action planning.

One limitation of the proposed method is that the trajectory generated by a diffusion model sometimes included actions that did not remove the shape.
Optimizing the trajectory based on the grinding resistance during processing is a further extension to reduce the task execution time.
We may study a method that adjusts the trajectory with a small amount of real data by black-box optimization based on a trajectory generated by a diffusion model.

\section{CONCLUSION}
We proposed novel sim-to-real transferable object shaping for grinding.
A novel technique that constrained a robot's action based on removal resistance theory reduced the sim-to-real gap. 
This technique enables data collection on a simplified geometric simulator and action planning with a diffusion model.
Experimental results show that our method enables a sim-to-real transfer and an action plan with a diffusion model quickly provides flexible long-horizon action sequences for different materials and various target shapes.
We found that the task execution times can be reduced in the grinding process, where shape observation is time-consuming.
\bibliographystyle{ieeetr}
\bibliography{reference}

\end{document}